\documentclass[10pt,twocolumn,letterpaper]{article}

\usepackage{3dv}
\usepackage{times}
\usepackage{epsfig}
\usepackage{graphicx}
\usepackage{amsmath}
\usepackage{amssymb}

\usepackage{subcaption}
\usepackage{color}
\usepackage[table]{xcolor}
\usepackage{bbold}
\usepackage{hhline}

\definecolor{col1}{RGB}{232, 161, 148}
\definecolor{col2}{RGB}{148, 187, 232}
\definecolor{col3}{RGB}{120, 200, 180}
\definecolor{Asectioncolor}{RGB}{255, 200, 200}
\definecolor{Bsectioncolor}{RGB}{255, 228, 196}
\definecolor{Csectioncolor}{RGB}{235, 255, 235}

\definecolor{3}{RGB}{240, 240, 240}
\definecolor{5}{RGB}{220, 220, 220}
\definecolor{7}{RGB}{200, 200, 200}

\usepackage{adjustbox}
\usepackage{chngpage}
\usepackage{multirow}
\usepackage{amssymb}
\usepackage{graphicx}
\usepackage[shortlabels]{enumitem}
\usepackage{array}
\usepackage{tabularx,booktabs}
\usepackage[labeled,resetlabels]{multibib} 
\newcites{S}{References} 
\usepackage{wrapfig} 
\usepackage{stackengine}
\usepackage{lipsum}
\usepackage[normalem]{ulem}

\usepackage{hhline}

\usepackage{float}
\usepackage{dblfloatfix}

\usepackage[pagebackref=false,breaklinks=true,letterpaper=true,colorlinks,bookmarks=false]{hyperref}

\threedvfinalcopy 

\ifthreedvfinal\fi

\begin{document}
\title{Attention meets Geometry: Geometry Guided Spatial-Temporal Attention for Consistent Self-Supervised Monocular Depth Estimation}
\author{
Patrick Ruhkamp$^{\ast}$ $^{1}$
\quad Daoyi Gao$^{\ast}$ $^{1}$
\quad Hanzhi Chen$^{\ast}$ $^{1}$
\quad Nassir Navab$^{1}$
\quad Benjamin Busam$^{1}$\\
$^{\ast}$ Equal contribution. Author ordering determined randomly. \qquad
$^1$ Technical University of Munich\qquad\\
{{\tt\small \{p.ruhkamp,...,b.busam\}@tum.de}
}}

\maketitle
\thispagestyle{empty}
\begin{abstract}
Inferring geometrically consistent dense 3D scenes across a tuple of temporally consecutive images remains challenging for self-supervised monocular depth prediction pipelines. This paper explores how the increasingly popular transformer architecture, together with novel regularized loss formulations, can improve depth consistency while preserving accuracy. We propose a spatial attention module that correlates coarse depth predictions to aggregate local geometric information. A novel temporal attention mechanism further processes the local geometric information in a global context across consecutive images. Additionally, we introduce geometric constraints between frames regularized by photometric cycle consistency. By combining our proposed regularization and the novel spatial-temporal-attention module we fully leverage both the geometric and appearance-based consistency across monocular frames. This yields geometrically meaningful attention and improves temporal depth stability and accuracy compared to previous methods.
\end{abstract}

\section{Introduction}
\label{sec:intro}
Improving the accuracy of self-supervised monocular depth prediction has been studied extensively over the past years~\cite{monodepth2,casser2019depth}.
However, predicting temporally and geometrically consistent depth over multiple consecutive frames in a self-supervised fashion is mostly unexplored.
Consistency is essential for many applications in 3D vision such as reconstruction~\cite{kinectfusion}, SLAM~\cite{Yang_2018_DVSO}, pose estimation~\cite{busam2017camera}, medical applications~\cite{busam2018markerless}, AR/MR~\cite{luo2020consistent}, computational photography~\cite{busam2019sterefo}, or autonomous driving~\cite{Geiger_2013}.

\begin{figure}[t]
      \centering
      \includegraphics[width=0.95\columnwidth]{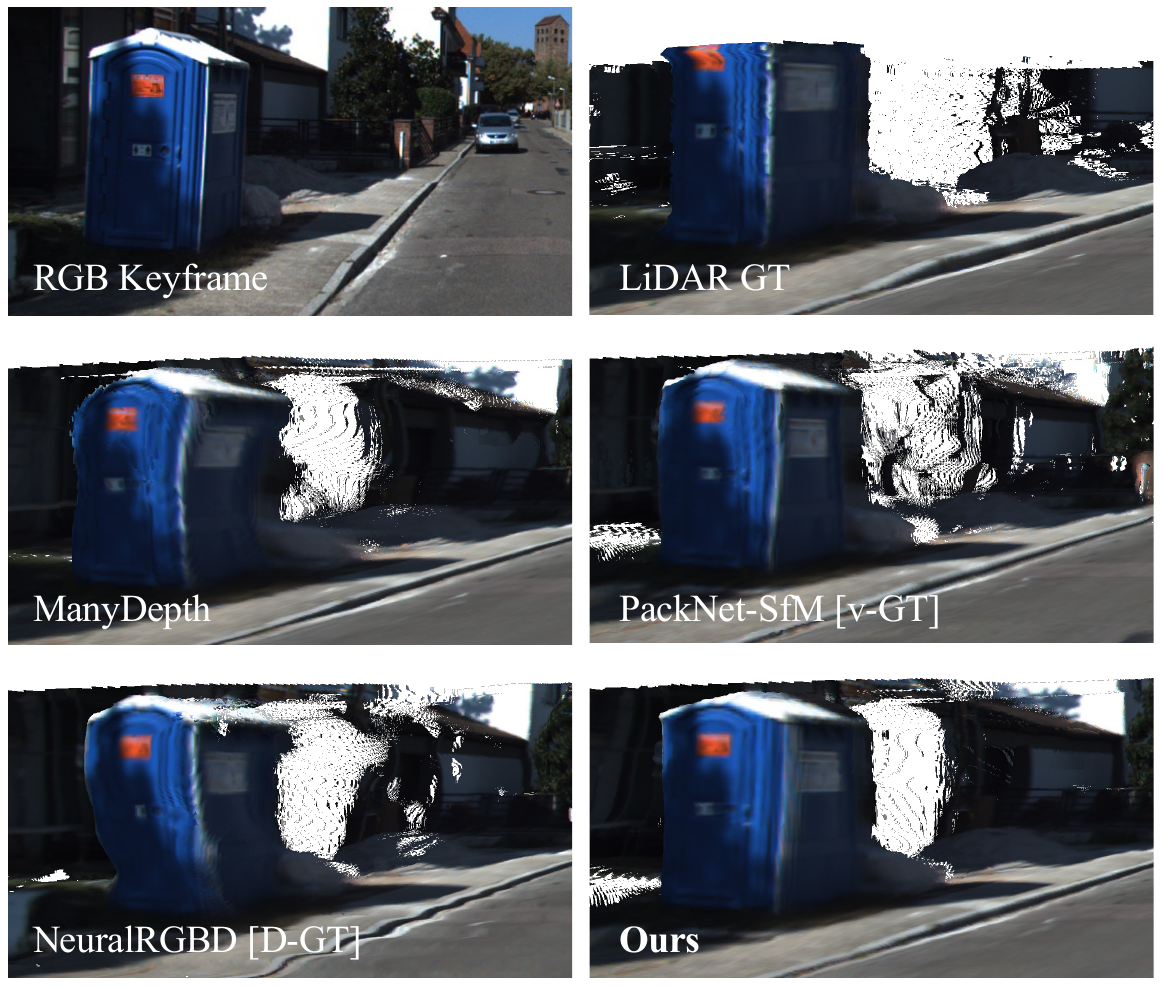}
      \caption{3D reconstruction from five consecutive depth predictions on Kitti~\cite{Geiger_2013}: Our method yields a higher quality reconstruction due to geometrically consistent depth predictions with high accuracy compared against SOTA methods in self-supervised (ManyDepth~\cite{watson2021temporal}), semi-supervised (PackNet-SfM~\cite{guizilini20203d} with pose velocity [v-GT]), and supervised (NeuralRGBD~\cite{liuneural} with depth [D-GT]) methods. Twisted boundaries due to pixel-wise misalignment and "flying pixels" are significantly reduced.}
      \label{teaser}
\end{figure}

\begin{figure*}[t]
\centering
\includegraphics[width=0.95\textwidth]{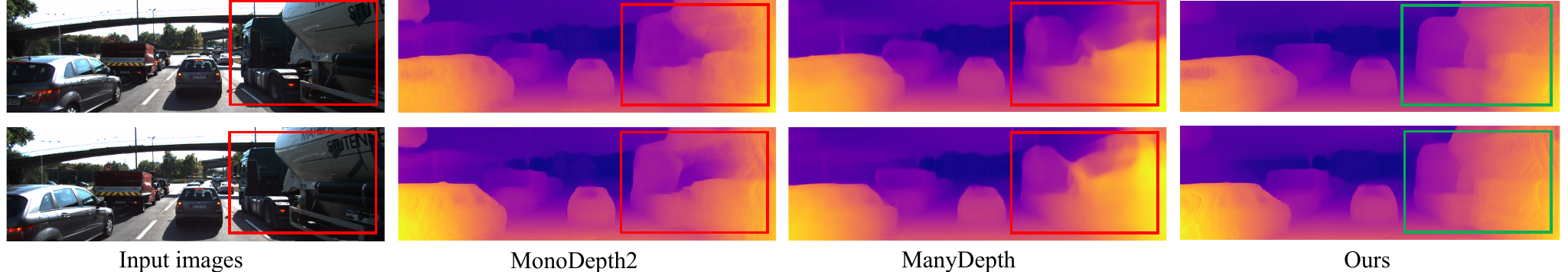}

\caption{Qualitative depth results: Strong baselines~\cite{monodepth2, watson2021temporal} in self-supervised depth prediction suffer from flickering effects between consecutive images. Our method estimates consistent depth across frames, even capable of handling large dynamically moving objects.}
\label{teaser_wide}
\end{figure*}
 
\paragraph{Consistent Depth Estimation}
Any downstream tasks may suffer from inconsistent dense depth predictions as for instance, inaccurate 3D object pose estimation in safety-critical applications for autonomous vehicles~\cite{Hu2021QD3DT,Hu2019Mono3DT} or RGB-D reconstruction~\cite{kinectfusion}.
Geometric consistency has long ago been examined in classical SfM and visual odometry, where usually computationally expensive local and global bundle adjustment aligns sparse triangulated points~\cite{mur2015orb,mur2017orb} to account for erroneous initial predictions. 
Some recent depth prediction pipelines try to enforce consistency~\cite{bian2019unsupervised} using additional ground truth signals such as velocity~\cite{guizilini20203d}, or take whole sequences into account to train with recurrent units~\cite{patil2020don}.

The evaluation of depth accuracy from monocular self-supervised methods usually employs median scaling against the depth ground truth to account for general scale ambiguity~\cite{monodepth2}.
As this is applied independently per image, the consistency of predictions is completely neglected in such metrics. 
Hence, the procedure is inadequate to capture the quality of the predictions for real-world scenarios and disregards pixel-wise variations in predictions across multiple frames.

We advocate paying more attention to the temporal consistency of depth predictions 
by proposing a new metric that allows to quantify this consistency. 
The qualitative 3D reconstruction~\cite{choi2015robust} from consecutive depth predictions in Fig.~\ref{teaser} gives an impression of the need for consistent depth predictions and justifies our motivation. 
While the very recently presented ManyDepth~\cite{watson2021temporal} currently achieves the best accuracy results (see Table~\ref{depth_results}), the inconsistent predictions result in noisy reconstructions of the scene. 
Our model does not only achieve good accuracy metrics but most importantly yields highly consistent depth predictions - even for dynamically moving objects as illustrated in Fig.~\ref{teaser_wide} - and subsequently improves reconstructions of the scene.

\paragraph{Contributions and Key Results}
Enforcing geometric consistency constraints usually negatively affect depth accuracy towards blurry edges and smooth depth discontinuities in self-supervised methods~\cite{bian2019unsupervised}.
Our \textbf{t}emporally \textbf{c}onsistent \textbf{depth} estimation pipeline, short \textbf{TC-Depth}\footnote{\url{https://daoyig.github.io/attention_meets_geometry/}}, 
enables explicit learning of temporally consistent features for depth prediction in a spatial-temporal attention module, together with geometric regularization, thus achieving high accuracy and unprecedented consistency. 
An extensive ablation study proves individual contributions on consistency and accuracy, and how our novel geometric constraint with photometric cycle consistency improves the attention mechanism significantly. 
To this end our contributions are:
\begin{enumerate}
    \item A novel \textbf{spatial attention} formulation which aggregates local geometric information.
    \item A \textbf{temporal attention} module across tuples of monocular frames which ensures global consistency.
    \item A novel \textbf{cycle consistency regularization} scheme for our \textbf{geometric guidance} of the spatial-temporal attention fusion of feature embeddings.
    \item A new \textbf{temporal consistency metric} (TCM) to quantify depth consistency across frames. 
\end{enumerate}
\section{Related Work}
\label{sec:related}
Recent pipelines~\cite{eigen2014depth,laina2016deeper,fu2018deep} pioneer the task of supervised monocular depth estimation with convolutional neural networks (CNNs). However, acquiring accurate ground-truth depth data remains difficult, especially for outdoor and large-scale scenes~\cite{Geiger_2013}. 
Scholars~\cite{xie2016deep3d,garg2016unsupervised} propose self-supervised learning approaches with photometric consistency losses by leveraging stereo imagery during training. 
Monodepth~\cite{Godard_2017} explores left-right consistencies in a fully differentiable pipeline which is also extended to the temporal domain in MonoDepth2~\cite{monodepth2}, where the complementary prediction of relative camera poses is necessary. 
While initial joint estimation of depth and pose~\cite{zhou2017unsupervised} falls short of accuracy compared to traditional methods, the robustness seems interesting~\cite{Babu_undemon_2018, zhang2019exploiting}.
The use of optical flow~\cite{Yin_2018,Yang_2019} greatly improves depth results in particular for moving objects in the scene where forward-backward consistency checks are used to automatically detect occlusions~\cite{Yin_2018,Janai2018ECCV,Wang_2018}.

\begin{figure*}[!ht]
      \centering
      \includegraphics[width=0.95\textwidth]{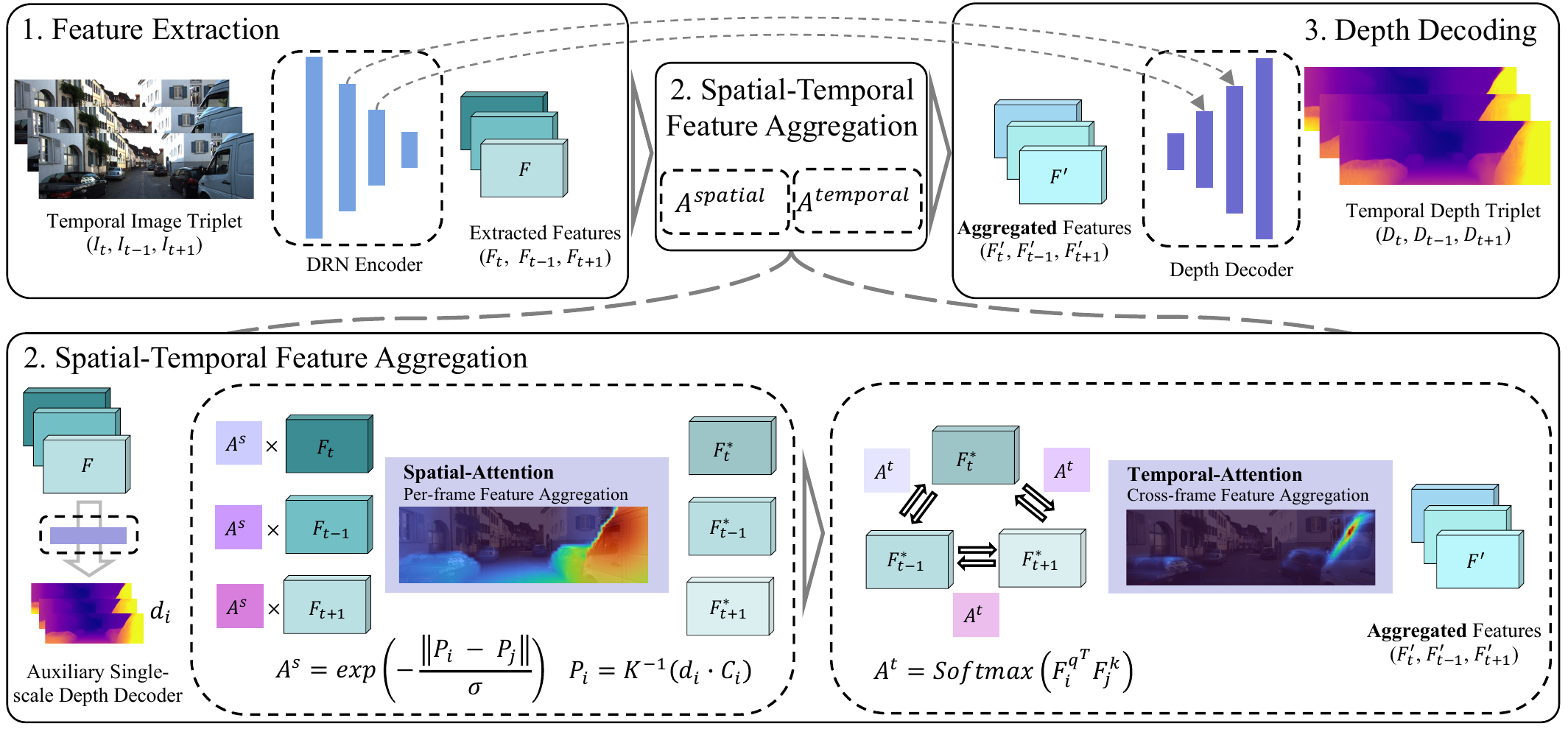}
      \caption{Pipeline Overview: 1. Image features are extracted with a dilated residual network (DRN) 2. An auxiliary low resolution depth map is predicted by a single-stage reference decoder and passed to the spatial attention module for local geometric correlation. The temporal attention aggregates the spatially-aware features globally across frames.  
      3. Aggregated features are decoded to final depth predictions with skip connections from the encoder. 
      }
      \label{pipeline}
\end{figure*}

\paragraph{Attention for Depth Estimation}
Self-attention mechanisms have shown impressive results in the field of natural language processing~\cite{vaswani2017attention} and are becoming increasingly popular in computer vision~\cite{zhao2020exploring,luo2020attention}.
While a trained set of traditional convolutions is applied independently to an image with fixed kernels during test time, self-attention constitutes a set of operations that adapt to the image and feature input. 
Huynh et al.~\cite{huynh2020guiding} propose a depth-attention volume to favour planar scene structures, well suited for indoor environments, while~\cite{sadek2020self} use attention gates in the decoding stage of depth estimation.
In~\cite{lee2021patch} patch-wise attention aggregates information of neighbouring features in the scene to predict dense depth in a supervised setting.
Also ~\cite{yang2021transformers} proposed the integration of transformers within a large architecture for highly accurate predictions, but only show applicability in a fully supervised setting. 
Johnston et al.~\cite{johnston2020self} pioneer the integration of transformers in self-supervised depth prediction for large outdoor scenes. They propose a self-attention mechanism on the feature embedding of input frames after a ResNet encoder and integrate a discrete disparity volume as depth decoder. Despite achieving good accuracy results, the naive self-attention seems non-expressive and not capable of aggregating meaningful feature correlation for the task of 3D scene regression.

\paragraph{Consistent Depth Estimation} 
Previous attempts to introduce metrics for depth consistency did not fully capture the geometric and temporal consistency.
\cite{zhang2019exploiting} tried to measure the structural similarity between two consecutive depth maps without aligning them spatially.
An optical-flow-based KLT tracker was borrowed in \cite{luo2020consistent} to measure Euclidean distances in 3D between photometrically corresponding points, thus being heavily influenced by the quality of the optical flow estimation.

Until now, self-supervised monocular depth pipelines mainly focus on maintaining a constant overall scale of depth predictions which also affects the auxiliary pose network being more suitable for odometry applications.
For this purpose, Bian et al.~\cite{bian2019unsupervised} propose a scale consistent depth and ego-motion approach by adding a depth consistency loss. 
This leads to reduced scale drift of inferred poses and depth but decreases depth accuracy. 
Zhao et al.~\cite{zhao2020towards} propose a method without direct regression of the 6-DOF camera transformation. They first estimate the optical flow between frames to sample correspondences for relative pose estimation via epipolar geometry. Consistency between triangulated points and the predicted depth ensures scale consistency.
MonoRec~\cite{wimbauer2020monorec} also focuses on visual odometry applications and achieves impressive results by building a photometric error cost volume to handle static and dynamic elements in the scene in a multi-view stereo setup. 
However, they employ additional supervision on dense stereo depth predictions and require a complex training scheme. 
Other works focus on static small-scale indoor scenes.
Luo et al.~\cite{luo2020consistent} use learning-based priors and test-time training for such scenarios. Their optimization method involves all pixels in a monocular video to achieve highly consistent small-scale reconstructions.
NeuralRGBD~\cite{liuneural} specifically focuses on consistency, by integrating multiple depth estimates from video sequences in a probability volume, thus aggregating consistent 3D scene information for indoor scene reconstruction in a supervised setting.
In order to exploit input image sequences further, Patil et al.~\cite{patil2020don} use recurrent units to learn more accurate depth predictions utilizing multiple frames in a self-supervised approach. The limitation of this approach, however, is the need for long sequences during both training and test time.
ManyDepth~\cite{watson2021temporal} proposes to utilize nearby frames of the monocular video sequence during inference time by proposing a cost volume which aggregates the encoded features of multiple frames. This approach is more efficient than previous test time refinement procedures~\cite{shu2020featdepth} and achieves highly accurate self-supervised depth predictions, yet relative poses between frames need to be predicted as well. 
In our analysis, however, the improved accuracy and the full utilization of multiple consecutive test frames do not necessarily manifest in temporally consistent depth predictions.

\section{Method}
\label{sec:method}
The goal of \textbf{TC-Depth} is to learn consistent and accurate depth from monocular image sequences in a self-supervised manner. 
We employ the widely used paradigm of regressing depth and relative camera poses jointly, by minimizing the image reconstruction loss after warping adjacent frames into a common central view via backwards warping with predicted dense depth and pose~\cite{monodepth2}.
We propose the network architecture as illustrated in Fig.~\ref{pipeline}. 
For pose regression, we employ the same strategy as previous methods~\cite{monodepth2, watson2021temporal} (not illustrated here).

We opt for a feature encoder with dilated convolutions~\cite{yu2017dilated} to align resolutions with the attention module in the bottleneck. 
The DRN-C-26 encoder is similar to a ResNet18 but with dilated strides and additional de-griding layers to remove checkerboard effects~\cite{yu2017dilated}. 
The feature embedding of the encoder is additionally given to an auxiliary single-scale depth decoder~\cite{johnston2020self, gonzalezbello2020forget} which produces a coarse initial depth prediction for the spatial-temporal attention module.
The attention mechanism is applied on the coarsest resolution at $24\times80$ which is $1/ 8th$ of the input resolution. 
Inspired by optical flow approaches, the temporal attention takes the encoded input features, together with the spatial attention, to aggregate temporally consistent scene content, before passing through the final depth decoder.

\subsection{Attention Module}
\begin{figure*}[!hbp]
      \centering
      \includegraphics[width=0.95\textwidth]{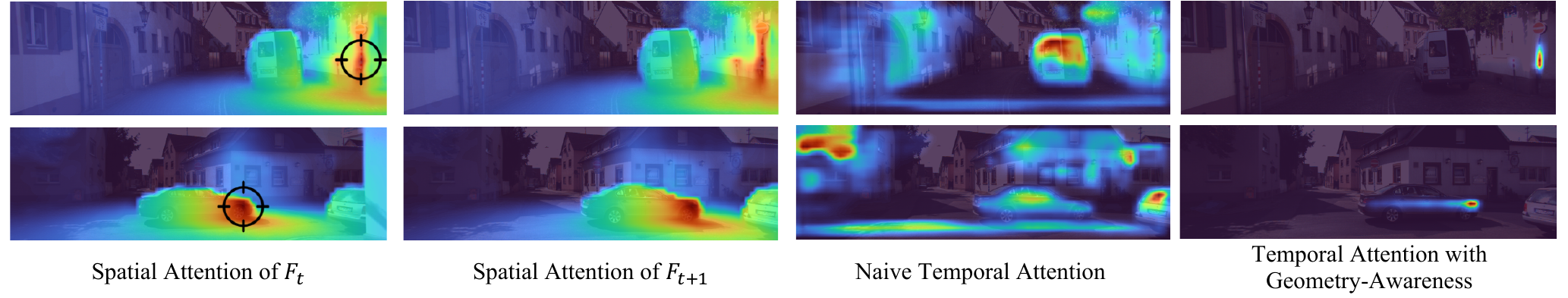}
      \caption{Spatial and temporal attention for a queried pixel (indicated by cross) between frames: The boundary of the spatial attention conforms to the scene structure. The appearance-based naive temporal attention seems unspecific. Our spatially-aware temporal attention focuses on visually similar features with geometric reasoning.
      }
      \label{spa_temp_att}
\end{figure*}
Convolutional neural networks are constrained by a receptive field which prohibits the correlation of features from spatially distant inputs.
Transformers have been proposed in NLP~\cite{vaswani2017attention} to correlate word items that share some semantic correlation but are far apart in the structure of a sentence. 
Similar approaches have been applied in computer vision~\cite{dosovitskiy2020image} where items are now pixels or some patch of pixels.
The inputs for the attention layer are usually named query (Q), key (K), and value (V).
Q retrieves information from V based on the attention weight. 
The attention is defined as:
\begin{equation}
\begin{aligned}
    \text{Attention}(Q,K,V) = \mathcal{A}(Q, K)V,
\end{aligned}
\end{equation} 
where $\mathcal{A}(\cdot)$ is a function that produces a similarity score as attention weight between feature embeddings for aggregation. 

Recent works~\cite{sun2021loftr} have shown that transformer models with self- and cross-attention can outperform fully convolution networks~\cite{li20dualrc} for the task of finding dense correspondences between image pairs.
Inspired by these findings we propose our spatial-temporal attention module.

\paragraph{Spatial-Attention Layer}
Self-attention as proposed in~\cite{johnston2020self} correlates information within the same image to attend to visually similar parts of the scene. 
The dot-product in the attention module can introduce some feature aggregation from geometrically distant parts in the 3D scene, which may not be desirable for the task of dense depth regression.

We propose explicit modelling of self-attention with 3D spatial awareness by exploiting a coarse predicted initial depth estimate.
Given known camera intrinsics $ \mathbf{K}$, a pair of coordinates $ \mathbf{C_i} = (u_i, v_i)$ and $\mathbf{C_j} = (u_j, v_j)$, together with their depth $d_i$ and $d_j$. We first back-project the two pixel coordinates to 3D space:
\begin{equation}
\begin{aligned}
    \mathbf{P}_i = \mathbf{K}^{-1}( {d_i} \cdot \mathbf{C}_i) ,\quad
    \mathbf{P}_j = \mathbf{K}^{-1}( {d_j} \cdot \mathbf{C}_j).
\end{aligned}
\end{equation}

Then we formulate the spatial-attention explicitly as: 
\begin{equation}
\begin{aligned}
    \mathcal{A}_{i,j}^{\textit{spatial}} = \exp \bigg( -\frac  { \| \mathbf{P}_i - \mathbf{P}_j  \|_2 }  {\sigma}\bigg),
\end{aligned}
\end{equation}
where $\mathbf{P}_i$, $\mathbf{P}_j$ can be treated as key and query, respectively. This can be interpreted as 3D positional encoding via 3D spatial correlation.
    
\paragraph{Temporal-Attention Layer}
Inspired by the correlation layer in optical flow~\cite{IMKDB17} and recent dense matching pipelines~\cite{sun2021loftr}, we formulate a novel temporal attention across frames by exploiting the temporal image sequence input of the self-supervised training scheme.

As a result, given a triplet of feature maps from consecutive image inputs, we can iteratively choose one of them as query and the rest as key features, and then acquire the key-query similarities using \text{Softmax}. Here we define $\mathbf{F}_i^q$ as query feature and $\mathbf{F}_{j}^k$ as key feature, and temporal-attention is formulated as:
\begin{equation}
\begin{aligned}
    \mathcal{A}_{i,j}^{\textit{temporal}} = \text{Softmax}_j ({\mathbf{F}_i^q} ^\top \mathbf{F}_{j}^k).
\end{aligned}
\end{equation} 

\paragraph{Spatial-Temporal Attention}
The unique formulation of our proposed spatial-temporal attention model can explicitly correlate geometrically meaningful and spatially coherent features - by first passing through the spatial attention - and at the same time provide temporal correlations across subsequent frames.
Fig.~\ref{spa_temp_att} visualizes the spatial and temporal attention individually for a queried pixel. The spatial attention aggregates geometrically consistent parts of the scene (notice large attention gradients towards the background at object edges). The appearance-based temporal attention correlates global information, which may be difficult and imprecise in a naive approach. With our additional geometric constraints (as later discussed in~\ref{geo_loss_motivation} and defined in Eq.~\ref{eqn:geoloss}) the attention is very focused and spatially coherent, as illustrated for two very challenging examples with thin structures and dynamic objects.

\subsection{Regularized Geometric Consistency}
\label{geo_loss_motivation}
\paragraph{Scale-invariant Consistent Depth Loss}
Constraining the absolute depth or disparity values between frames after projecting into the same camera view would either shrink or enlarge the overall scene depth-scale. 
Scale-invariant formulations have been proposed~\cite{zhao2020towards, bian2019unsupervised, luo2020consistent}, but they do not provide strong gradients for depth values that exhibit small alignment errors.
Therefore, we adopt a formulation of~\cite{kopf2021robust}, with additional regularization as detailed in Eq.~\ref{eqn:geoloss}, to constraint depth predictions to be consistent between frames.

\begin{figure}[!t]
      \centering
      \includegraphics[width=\columnwidth]{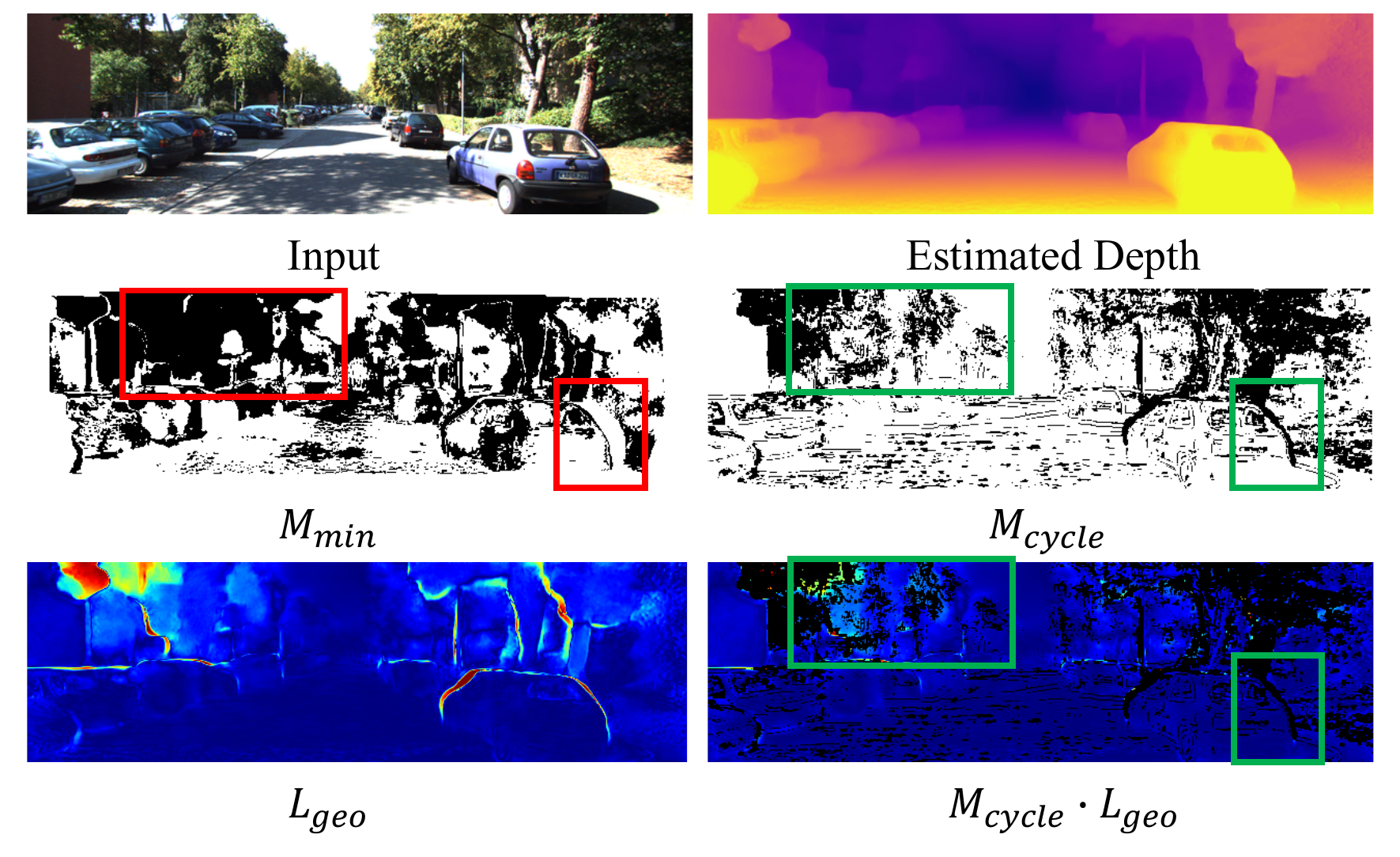}
      \caption{Visualization of occlusion handling for $\mathcal{L}_{\text{geo}}$ with $\mathcal{M}_{\text{cycle}}$ as opposed to pixel-wise minimum: The $\mathcal{M}_{\text{min}}$ cannot robustly account for all occlusions and falsely masks out large image regions. $\mathcal{M}_{\text{cycle}}$ can instead model such cases, resulting in better gradients for training with $\mathcal{M}_{\text{cycle}}\cdot \mathcal{L}_{\text{geo}}$.}
      \label{occlusion_vis}
\end{figure}

\paragraph{Cycle-Mask from Photometric Consistency}
Aggregating the pixel-wise mean geometric loss over different views violates the scene structure as occluded regions would contribute to the loss computation, resulting in blurry edges and low depth accuracy~\cite{bian2019unsupervised}.
The pixel-wise minimum depth error was already proposed to avoid this issue~\cite{zhan2019dfvo, gao2020attentional}.
However, quantitative and qualitative evaluations show that this strategy, while mostly solving the issue of occluded regions, often also excludes major regions of the scene. These can be regions with large inconsistency due to imprecise transformation of adjacent depth maps.
The minimum operator can mask out large regions of the scene (see Fig.~\ref{occlusion_vis}), which harms the training signal.
Instead, we propose a novel masking scheme by exploiting the assumption of photo-consistency.
For this purpose, the central target image $I_{t}$ is projectively transformed to the view of the adjacent source frame $I_{t\rightarrow s}$ and then transformed back again $I_{t \rightarrow s \rightarrow t}$. 
Our cycle-masking can be formulated as:
\begin{equation}
\begin{aligned}
    \mathcal{M}_{\text{cycle}} = \big[ {E}_\text{pe}(I_t, I_{t \rightarrow s \rightarrow t}) < \gamma \big], 
\label{eqn:cycle_mask}
\end{aligned}
\end{equation} 
where $[\cdot]$ is the Iverson bracket and ${E}_\text{pe}$ is the photometric error as defined in Eq.~\ref{eqn:photo}. 
We set an adaptive threshold $\gamma$ as the $70\%$ percentile of the photometric error among all pixels of $I_s$ for binarization of $\mathcal{M}_{\text{cycle}}$. 
With our cycle-masking, we can successfully rule out occluded regions while preserving most of the non-occluded regions for more exhaustive geometric consistency checking as illustrated in Fig.~\ref{occlusion_vis}.

\subsection{Loss Formulation}
Our model is trained with a set of loss terms based on content-based image reconstruction and geometric properties of our depth map. It reads:
\begin{equation}
    \mathcal{L} =  \mathcal{L}_{\text{photo}}
    + \lambda_{\text{s}} \mathcal{L}_{\text{s}}
    + \lambda_{\text{geo}} \mathcal{L}_{\text{geo}} 
    + \lambda_{\text{m}} \mathcal{L}_{\text{m}}
    + \mathcal{L}_{\text{ref}},
\end{equation}
where $\mathcal{L}_{\text{photo}}$ and $\mathcal{L}_{\text{s}}$ follow previous established methods~\cite{monodepth2, watson2021temporal} and will therefore be only briefly mentioned. We detail all other parts hereafter.

\paragraph{Motion Consistency Loss $\mathcal{L}_{\text{m}}$}
Inspired by the knowledge distillation strategy from \cite{pilzer2019refine}, we train a simplified self-supervised depth prediction network (MonoDepth2~\cite{monodepth2} in Table~\ref{depth_results}) alongside as weak teacher. 
Following~\cite{watson2021temporal}, we define a mask where large differences between our prediction $D_t$ and the teacher $\hat{D_t}$ may indicate moving objects, which is also utilized for the photometric loss later, as
\begin{align}
\mathcal{M}_{\text{m}} = \max
    \Big(
        \frac{D_t - \hat{D_t}}{\hat{D_t}}, \frac{\hat{D_t} - D_t}{D_t}
    \Big) < 0.6.
\end{align}
This yields our motion consistency loss term to help the student to learn from the weak teacher as
\begin{align}
\mathcal{L}_{\text{m}} = (1-\mathcal{M}_{\text{m}}) \cdot \, \Vert D_t - \hat{D_t}\Vert_{1}.
\end{align}

\paragraph{Photometric Loss $\mathcal{L}_{\text{photo}}$} 
The photometric reconstruction error~\cite{monodepth2, watson2021temporal} between image $I_x$ and $I_y$ given by
\begin{align}
    {E_{\text{pe}}}(I_x,I_y) = \alpha\tfrac{1-\text{SSIM}(I_x, I_y)}{2} + (1-\alpha) \left \Vert I_x-I_y \right \Vert_{1}
\label{eqn:photo}
\end{align}
is computed between the target frame $I_t$ and each source frame $I_s$ with $s \in S$ and the pixel-wise minimum error is retrieved. 
An auto-mask accounts for objects moving with the same velocity and direction as the camera-ego motion
\begin{align}
    \mathcal{M}_{\text{auto}} = 
    \big[  \min_{s \in S}  {{E_{\text{pe}}}(I_t,I_{s \rightarrow t})}  < 
             \min_{s \in S}  {E_{\text{pe}}} (I_t,I_s) \big].
\end{align}
$\mathcal{L}_{\text{photo}}$ is finally defined over $S \in \{t-1, t+1 \}$ as
\begin{align}
    \mathcal{L}_{\text{photo}} = \mathcal{M}_{\text{m}} \cdot \mathcal{M}_{\text{auto}} \cdot \min_{s \in S}  {E_{\text{pe}}}(I_t,I_{s \rightarrow t}) .
\end{align}

\paragraph{Edge-aware Smoothness Loss $\mathcal{L}_{\text{s}}$}
The edge-aware smoothness is applied as in previous works~\cite{Godard_2017, monodepth2} to encourage locally smooth depth estimations with the mean-normalized inverse depth $\overline{d_{t}}$ as
\begin{align}
\mathcal{L}_{\text{s}} = \left|\partial_{x}\overline{d_{t}}\right|e^{-\left|\partial_{x}I_{t}\right|} + \left|\partial_{y}\overline{d_{t}}\right|e^{-\left|\partial_{y}I_{t}\right|}.
\end{align}

\paragraph{Geometric Loss $\mathcal{L}_{\text{geo}}$}
As motivated in Sec.~\ref{geo_loss_motivation}, 
we design a geometric loss to encourage consistent depth predictions between frames that not only alleviates the problem of penalizing the scale of the depth prediction, but also utilizes the cycle consistency (Eq.~\ref{eqn:cycle_mask}) to handle occlusions with
\begin{align}
\label{eqn:geoloss}
    \mathcal{L}_{\text{geo}} = \mathcal{M}_{\text{m}} \cdot \mathcal{M}_{\text{auto}} \cdot \mathcal{M}_{\text{cycle}} \cdot \Big( 1 - \frac{\min(D_{s \rightarrow t},  D^{\prime}_{t})}{\max(D_{s \rightarrow t}, D^{\prime}_{t})} \Big),
\end{align}
where the $D_{s \rightarrow t}$ is the depth map warped from the adjacent source frame to the target frame and $D^{\prime}_{t}$ is the interpolated target depth map~\cite{bian2019unsupervised, gao2020attentional}.

\paragraph{Reference Loss $\mathcal{L}_{\text{ref}}$}
To train the single-stage auxiliary depth decoder $D_{ref}$ for spatial attention acquisition, we minimize its difference against the (detached) final depth prediction of our full pipeline $D_{t}$:
\begin{align}
    \mathcal{L}_{\text{ref}} =\left \Vert D_{t} - D_{ref}\right \Vert_{1}.
\end{align}

\section{Temporal Consistency Metric (TCM)}
\label{tcm}
We propose to measure the consistency directly on the predicted depth output after aligning a number of $k$ frames in 3D via projective transformation, where $k$ is chosen to be in $\{3,5,7\}$ (longer sequences usually do not have enough visual overlap for outdoor driving scenes).
To transform all predictions from $I_s$ in a common reference frame of $I_t$, we use the ground-truth depth and pose. 
Monocular methods (with scale-ambiguity) are first aligned with the same median scaling ratio. 
Our temporal consistency metric (TCM) measures the track difference between estimated pixel-wise depth and GT across multiple frames.
A visual impression of TCM is given in Fig.~\ref{fig:tcm}. 
To account for errors in the interpolated ground-truth LiDAR and moving objects, we filter out 20\% of the largest outliers for a fair comparison.
In Sec.~\ref{sec:depth_consistency} more details on TCM results are given. 

\setcounter{figure}{5}
\begin{figure}[!t]
      \centering
      \includegraphics[width=0.95\columnwidth]{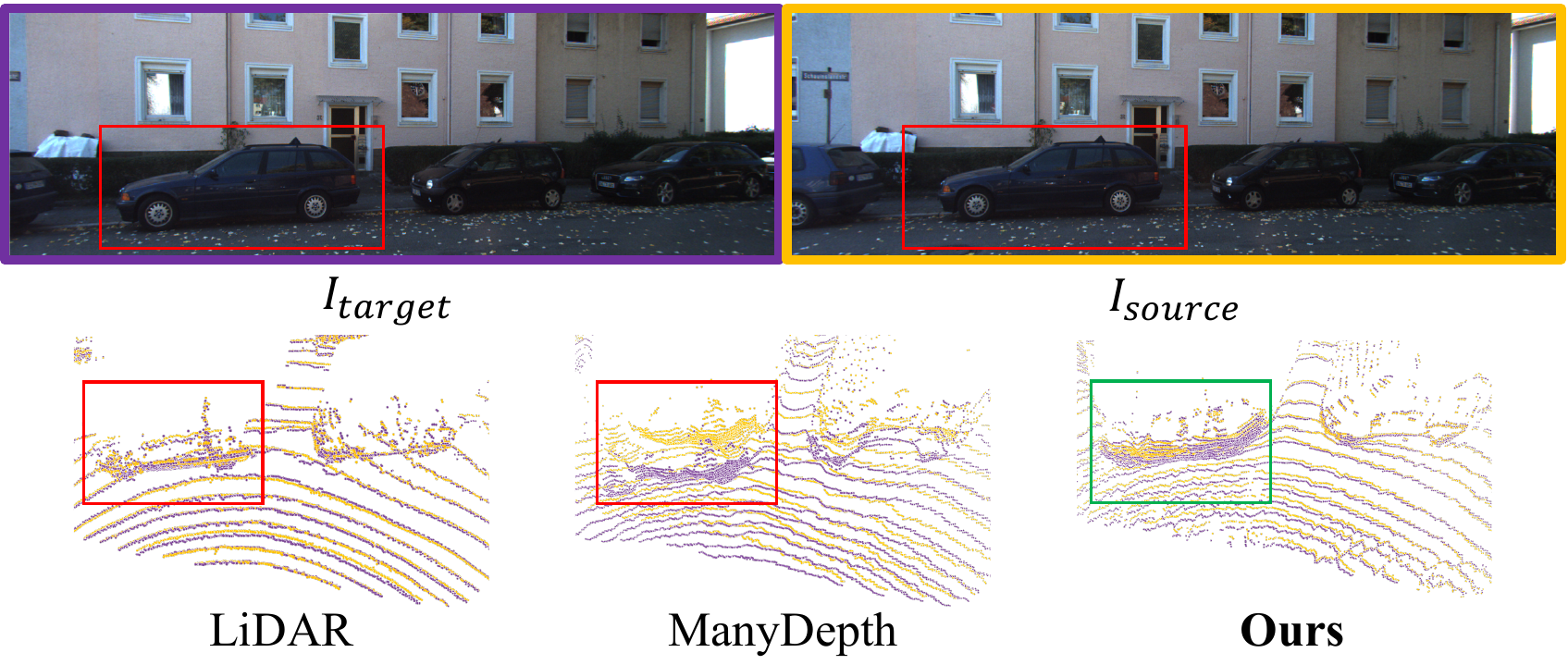}
      \caption{Visualization of TCM: Consecutive depth predictions are aligned in 3D and evaluated pixel-wise across multiple frames. The car shows significantly better alignment between frames for our method.}
      \label{fig:tcm}
\end{figure}
\setcounter{figure}{7}

\section{Experiments}
\label{sec:experiments}
We evaluate our model against recent SOTA quantitatively on temporal consistency with our proposed TCM and on well established depth accuracy metrics~\cite{monodepth2}.
We follow previous works on self-supervised depth estimation~\cite{monodepth2, watson2021temporal} and conduct extensive experiments on the Eigen split~\cite{eigen2014depth} of the Kitti dataset~\cite{Geiger_2013} and also report results on Cityscapes~\cite{Cordts_2016}. 
TCM is computed on a test split from the Kitti odometry data (details in Supplementary Material). Cityscapes is neglected for TCM as too many moving objects prevent reliable evaluation.
For inference we use an image triplet as indicated in Fig.~\ref{pipeline} similar to ManyDepth~\cite{watson2021temporal} where consecutive images are used. 
Different from ManyDepth~\cite{watson2021temporal}, our method does not need to predict relative poses between adjacent frames for depth inference. 
In our extensive ablation studies we observe that our model is not majorly influenced by the encoder with dilated convolutions, compared to the standard ResNet as in~\cite{monodepth2}, for consistent depth predictions, likewise for depth accuracy results.
Additional qualitative results give an impression of the superiority of our model, especially on temporally consistent 3D reconstructions from consecutive depth predictions (see Figs.~\ref{teaser},~\ref{teaser_wide},~\ref{reconstructions}).
Here we focus on the most important findings. For implementation details, and additional quantitative and qualitative results, we refer the interested reader to the supplementary material and video on our project website. 

\subsection{Depth Consistency}
\label{sec:depth_consistency}
Table~\ref{results_tcm_abs} summarizes the results on relative TCM for depth consistency over an increasing number of test frames. 
\textbf{TC-Depth} outperforms strong self-supervised baselines such as MonoDepth2~\cite{monodepth2}, SC-SfMLearner~\cite{bian2019unsupervised} which focuses on temporal consistency, and ManyDepth~\cite{watson2021temporal} which specifically utilizes adjacent frames during inference. Our method is also superior to the semi-supervised method of PackNet-SfM~\cite{guizilini20203d} and even NeuralRGBD~\cite{liuneural} with full supervision and the use of GT poses during testing. 

\setcounter{figure}{6}
\begin{table}[!t]
    \centering
    \footnotesize
    \resizebox{\columnwidth}{!}{
    \begin{tabular}{|l c c cc c cc c c|}
        \hline
        Method  & \multicolumn{3}{|c}{\cellcolor{col3}Abs Err} & \multicolumn{3}{|c}{\cellcolor{col3}Sq Err} & \multicolumn{3}{|c|}{\cellcolor{col3}RMSE} \\
        \hline
        \multicolumn{1}{|l}{\# Test Frames} & \multicolumn{1}{|c}{\cellcolor{3}3}  & \cellcolor{5}5  & \cellcolor{7}7 & \multicolumn{1}{|c}{\cellcolor{3}3}  & \cellcolor{5}5  & \cellcolor{7}7  & \multicolumn{1}{|c}{\cellcolor{3}3}  & \cellcolor{5}5  & \cellcolor{7}7 \\
        \hline
        \hline
        ManyDepth~\cite{watson2021temporal} &  \multicolumn{1}{|c}{\cellcolor{3}0.204}  & \cellcolor{5}0.260  & \cellcolor{7}0.307  & 
                                                \multicolumn{1}{|c}{\cellcolor{3}0.087}  & \cellcolor{5}0.147  & \cellcolor{7}0.206  & 
                                                \multicolumn{1}{|c}{\cellcolor{3}0.256}  & \cellcolor{5}0.319  & \cellcolor{7}0.373 \\

        MonoDepth2~\cite{monodepth2}   &        \multicolumn{1}{|c}{\cellcolor{3}0.137}  & \cellcolor{5}0.177  & \cellcolor{7}0.215  & 
                                                \multicolumn{1}{|c}{\cellcolor{3}0.039}  & \cellcolor{5}0.068  & \cellcolor{7}0.104  & 
                                                \multicolumn{1}{|c}{\cellcolor{3}0.176}  & \cellcolor{5}0.223  & \cellcolor{7}0.268 \\
                                                
        SC-SfMLearner~\cite{bian2019unsupervised} &       \multicolumn{1}{|c}{\cellcolor{3}0.126}  & \cellcolor{5}0.170  & \cellcolor{7}0.211  & 
                                                \multicolumn{1}{|c}{\cellcolor{3}0.032}  & \cellcolor{5}0.062  & \cellcolor{7}0.099  & 
                                                \multicolumn{1}{|c}{\cellcolor{3}0.159}  & \cellcolor{5}0.210  & \cellcolor{7}0.259 \\

        PackNet-SfM~\cite{guizilini20203d} &   \multicolumn{1}{|c}{\cellcolor{3}0.141}  & \cellcolor{5}0.196  & \cellcolor{7}0.247  & 
                                                \multicolumn{1}{|c}{\cellcolor{3}0.044}  & \cellcolor{5}0.090  & \cellcolor{7}0.147  & 
                                                \multicolumn{1}{|c}{\cellcolor{3}0.177}  & \cellcolor{5}0.240  & \cellcolor{7}0.299 \\
        
        PackNet-SfM~\cite{guizilini20203d}$^{*}$  & \multicolumn{1}{|c}{\cellcolor{3}0.118}  & \cellcolor{5}0.154  & \cellcolor{7}0.190  & 
                                                \multicolumn{1}{|c}{\cellcolor{3}0.030}  & \cellcolor{5}0.052  & \cellcolor{7}0.083  & 
                                                \multicolumn{1}{|c}{\cellcolor{3}0.154}  & \cellcolor{5}0.197  & \cellcolor{7}0.240 \\  
        
        NeuralRGBD~\cite{liuneural}$^{**}$   &       \multicolumn{1}{|c}{\cellcolor{3}0.116}  & \cellcolor{5}0.148  & \cellcolor{7}0.179  & 
                                                \multicolumn{1}{|c}{\cellcolor{3}0.024}  & \cellcolor{5}0.044  & \cellcolor{7}0.066  & 
                                                \multicolumn{1}{|c}{\cellcolor{3}0.147}  & \cellcolor{5}0.186  & \cellcolor{7}0.222 \\

        \textbf{Ours DRN-C-26}   &                      \multicolumn{1}{|c}{\cellcolor{3}\textbf{0.079}}  & \cellcolor{5}\textbf{0.111}  & \cellcolor{7}\textbf{0.147}  & 
                                                \multicolumn{1}{|c}{\cellcolor{3}\textbf{0.011}}  & \cellcolor{5}\textbf{0.025}  & \cellcolor{7}\textbf{0.047}  & 
                                                \multicolumn{1}{|c}{\cellcolor{3}\textbf{0.099}}  & \cellcolor{5}\textbf{0.139}  & \cellcolor{7}\textbf{0.184} \\
        
        \hline
        \textbf{Ours DRN-D-54}   &              \multicolumn{1}{|c}{\cellcolor{3}\textbf{0.076}}  & \cellcolor{5}\textbf{0.106}  & \cellcolor{7}\textbf{0.138}  & 
                                              \multicolumn{1}{|c}{ \cellcolor{3}\textbf{0.010}}  & \cellcolor{5}\textbf{0.022}  & \cellcolor{7}\textbf{0.041}  & 
                                                \multicolumn{1}{|c}{\cellcolor{3}\textbf{0.095}}  & \cellcolor{5}\textbf{0.131}  & \cellcolor{7}\textbf{0.172} \\             
        \hline                                        
        \end{tabular}
        }
        \captionof{table}{Temporal consistency metric (TCM) for increasing number of test frames [3, 5, 7]. $^{*}$:~semi-supervision with velocity. $^{**}$:~supervision with GT depth and inference with GT pose. Our self-supervised model improves TCM about 60\% across all metrics compared to strong baselines that leverage temporal frames~\cite{watson2021temporal}. It even outperforms semi-supervised~\cite{guizilini20203d} and fully supervised~\cite{liuneural} pipelines that aim to estimate temporally coherent depth.}
        \label{results_tcm_abs}
  \end{table}
  
\subsection{Depth Accuracy}
Table~\ref{depth_results} shows the depth accuracy results. 
Our model performs significantly better than comparable self-supervised models such as MonoDepth2~\cite{monodepth2} and can also yield better results than models with larger backbones (FeatDepth~\cite{shu2020featdepth}), models trained with consistency constraints (SC-SfMLearner~\cite{bian2019unsupervised}) or semi-supervised metods (PackNet-SfM~\cite{guizilini20203d}). 
We also adopt the test time refinement scheme (TTR in Table~\ref{depth_results}) of~\cite{mccraith2020monocular}, for which our method actually outperforms ManyDepth~\cite{watson2021temporal}.
Our method also achieves the best accuracy on the challenging Cityscapes dataset~\cite{Cordts_2016}. 

\begin{table}[!b]
\footnotesize
\begin{center}
\resizebox{\columnwidth}{!}{
\begin{tabular}{|lc|c|c|c|c|c|}
\hline
 Method && \cellcolor{col1} Abs Rel &\cellcolor{col1} Sq Rel &\cellcolor{col1} RMSE &\cellcolor{col2} $\sigma<1.25$ & \cellcolor{col2} $\sigma<1.25^{3}$ \\
\hline
\hline
Monodepth2~\cite{monodepth2}	&  	&	0.115	&	0.903  & 4.863	&	0.877 & 0.981\\
SC-SfMLearner~\cite{bian2019unsupervised} $\dagger$  &&  0.119   &   0.857  & 4.950  &   0.863&0.981 \\
TrianFlow~\cite{zhao2020towards}    &&  0.113   &   \textbf{0.704}   &  4.581 &  0.871&\textbf{0.984}\\
PackNet-SfM\cite{guizilini20203d}$^{*} $  &&   0.111   &  0.829 & 4.788 &  0.864 & 0.980  \\
FeatDepth\cite{shu2020featdepth} $\ddagger$   &&   0.109  &   0.923  & 4.819 &   0.886&0.981 \\
ManyDepth~\cite{watson2021temporal}     &&  \textbf{0.098} & \dotuline{0.770} &  \textbf{4.459} & \textbf{0.900}& \underline{0.983}\\
\textbf{Ours   (DRN-C-26)}      && \dotuline{0.106}  &   \dotuline{0.770} & \dotuline{4.558} &  \dotuline{0.890}& \underline{0.983}\\
\textbf{Ours (DRN-D-54)}   &&   \underline{0.103}  &    \underline{0.746} &   \underline{4.483} & \underline{0.894} & \underline{0.983}\\
\hline
\hline
ManyDepth\cite{watson2021temporal} & \multirow{2}{*}{\rotatebox[origin=c]{90}{TTR}}    &   \underline{0.090}  &   \underline{0.713}  & \underline{4.137} &   \underline{0.914}& \textbf{0.997} \\
\textbf{Ours   (DRN-C-26) }   &  &  \textbf{0.082} & \textbf{0.667} & \textbf{4.104} & \textbf{0.921} & \textbf{0.997}\\
\hline
\hline
Monodepth2~\cite{monodepth2} &CS&	\dotuline{0.129}	&	\dotuline{1.569}  & \dotuline{6.876}	&	\dotuline{0.849} & \dotuline{0.983} \\
ManyDepth~\cite{watson2021temporal}&CS&  \underline{0.114}   &   \underline{1.193}  & \underline{6.223}  &   \textbf{0.875} & \underline{0.989} \\
\textbf{Ours   (DRN-C-26)}    &CS&  \textbf{0.110}   &   \textbf{0.958}   &  \textbf{5.820} &  \underline{0.867} & \textbf{0.991}\\
\hline
\end{tabular}
}
\caption{Accuracy results on Kitti Eigen test split~\cite{eigen2014depth} for self-supervised monocular methods ($^{*}$ indicate semi-supervision). Middle: with test time refinement (TTR)~\cite{shu2020featdepth}. Bottom: Cityscape dataset~\cite{Cordts_2016}. 
$\dagger$: new results from GitHub; $\ddagger$: retrained results with standard image size for fair comparison. We highlight \textbf{best}; \underline{2nd best}; \dotuline{3rd best} results.
}
\label{depth_results}
\end{center}
\end{table}

\begin{table*}[!h]
\footnotesize
\begin{center}
\resizebox{\textwidth}{!}{
\begin{tabular}{|c|c|c|c|c|c|c|c|c|c|c|c||c|c|c|}
\hhline{------------||---}
\multirow{3}{*}{Model} & \multicolumn{4}{c|}{Ablations} & \multicolumn{7}{c||}{Accuracy} & \multicolumn{3}{c|}{TCM (3 Frames)} \\
\hhline{~-----------||---}
&\multirow{1}{*}{$\mathcal{L}_{\text{geo}}$} & \multirow{1}{*}{$\mathcal{M}_{\text{cycle}}$} & \multirow{1}{*}{\parbox{1.2cm}{\centering Attention}} &\multirow{1}{*}{$\mathcal{L}_{\text{m}}$}& \multirow{-1}{*}{\cellcolor{col1} Abs Rel} & \multirow{-1}{*}{\cellcolor{col1} Sq Rel} & \multirow{-1}{*}{\cellcolor{col1} RMSE} & \multirow{-1}{*}{\cellcolor{col1} RMSE log} & \multirow{-1}{*}{\cellcolor{col2} $\sigma<1.25$} & \multirow{-1}{*}{\cellcolor{col2} $\sigma < 1.25^{2}$} & \multirow{-1}{*}{\cellcolor{col2} $\sigma < 1.25^{3}$}  & \multirow{-1}{*}{\cellcolor{col3} Abs Err} & \multirow{-1}{*}{\cellcolor{col3} Sq Err} & \multirow{-1}{*}{\cellcolor{col3} RMSE}\\
\hhline{============||===}
MD2~\cite{monodepth2}& & & & &	                        0.115	&	0.903	&	4.863	&	0.193	&	0.877	&	0.959	&	0.981 &  0.137  &   0.039  &   0.176	\\
\hhline{------------||---}
\parbox[t][][c]{2mm}{\multirow{7}{*}{\rotatebox[origin=c]{90}{\centering DRN-C-26}}}
&& & & &                                                 0.115  &   1.027  &   5.004  &   0.197  &   0.879  &   0.958  &   0.979 & 0.136  &   0.039  &   0.173  \\
&\checkmark & & & &                                      0.113  &   0.904  &   4.773  &   0.193  &   0.877  &   0.959  &   0.980 & 0.124  &   0.032  &   0.157  \\
&\checkmark & \checkmark & & &                           0.111  &   0.878  &   4.761  &   0.190  &   0.882  &   0.961  &   0.981 & 0.113  &   0.026  &   0.141  \\
&&& S-A &&   0.113  &   0.958  &   4.861  &   0.192  &   0.882  &   0.960  &   0.980   & 0.134  &   0.038  &   0.172  \\
&&&T-A & &  0.116  &   1.028  &   5.024  &   0.197  &   0.879  &   0.957  &   0.979  &   0.133  &   0.037  &   0.171   \\
&& & ST-A & &                                      0.112  &   0.974  &   4.921  &   0.194  &   0.882  &   0.960  &   0.980 & 0.130  &   0.035  &   0.165  \\
&& & & \checkmark &                                      0.112  &   0.840  &   4.683  &   0.189  &   0.880  &   0.961  &   0.982 & 0.132  &   0.036  &   0.169  \\
&\checkmark & \checkmark &  ST-A & &                0.108  &   0.819  &   4.655  &   0.186  &   0.886  &   0.962  &   0.982 & 0.105  &   0.022  &   0.133  \\
&\checkmark & \checkmark & ST-A & \checkmark &     \textbf{0.106}  &   \textbf{0.770}  &   \textbf{4.558}  &   \textbf{0.182}  &   \textbf{0.890}  &   \textbf{0.964}  &   \textbf{0.983} & \textbf{0.079}  &   \textbf{0.011}  &   \textbf{0.099} \\
\hhline{------------||---}
DRN-D-54 &\checkmark & \checkmark &  ST-A & \checkmark &   0.103  &   0.746  &   4.483  &   0.180  &   0.894  &   0.965  &   0.983 & 0.076  &   0.010  &   0.095  \\
\hhline{------------||---}
\end{tabular}
}
\caption{Ablation study on depth accuracy and depth consistency (TCM). The consecutive activation of individual pipeline components as indicated all positively influence the overall performance of our method.
}
\label{ablations}
\end{center}
\end{table*}

\subsection{Ablation Study}

\begin{figure}[!t]
      \centering
      \includegraphics[width=0.95\columnwidth]{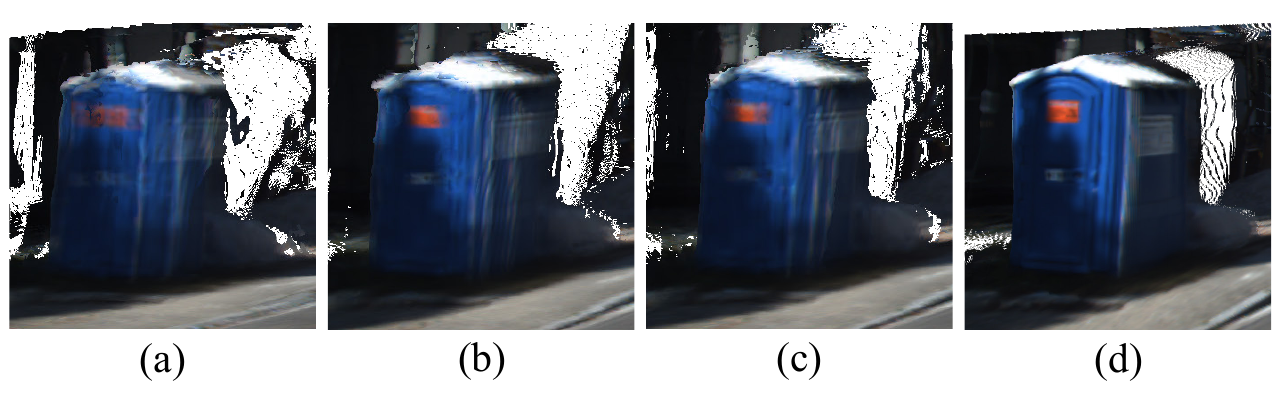}
      \caption{
      The baseline model without our contributions (a) suffers from strong ghosting effects due to erroneous pixel-wise misalignment. Constraining with $\mathcal{M}_{\text{cycle}}\cdot\mathcal{L}_{\text{geo}}$ (b) or applying spatial-temporal attention (c) mitigate such issue to a large extent. Our full model (d) yields the result with the highest quality.
      }
      \label{rec_ablation}
\end{figure}

\setcounter{figure}{9}
\begin{figure*}[!hb]
      \centering
      \includegraphics[width=0.95\textwidth]{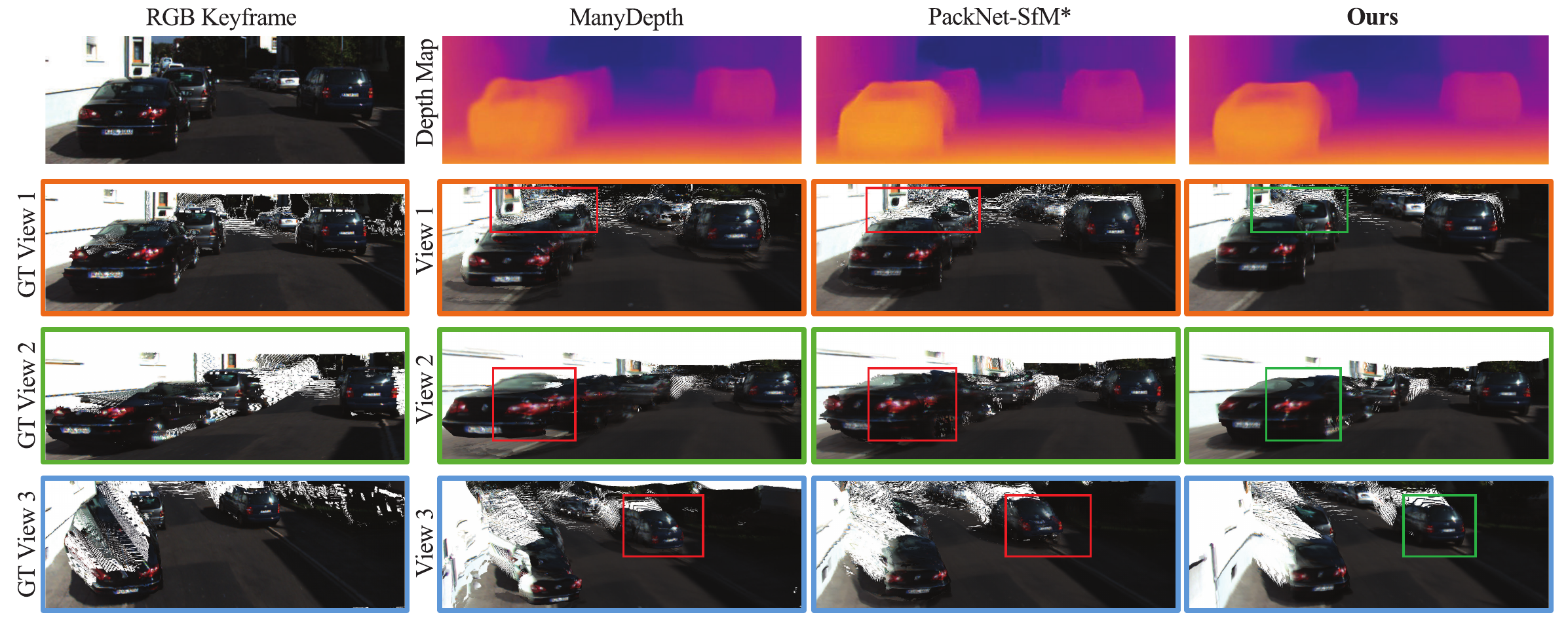}
      \caption{Qualitative reconstruction results from five consecutive depth predictions. Both, ManyDepth~\cite{watson2021temporal} and PackNet-SfM$^{*}$~\cite{guizilini20203d} with velocity semi-supervision, suffer from "flying pixels" (View 1), ghosting effects (View 2), and deformed objects (View 3), due to temporal inconsistencies. This is not directly apparent in a single frame depth prediction, but unfold when changing the viewpoint. Our method mitigates these artifacts to a large extend.
      }
      \label{reconstructions}
\end{figure*}
To quantitatively evaluate the influence of each sub-module of our pipeline, we perform an extensive ablation study and report TCM results and depth accuracy as before in Table~\ref{ablations}. 
The choice of the backbone (ResNet18 in MD2~\cite{monodepth2} against DRN-C-26 in our baseline) has only a marginal effect on accuracy and TCM. 

The ablation study reveals that the spatial-temporal attention (ST-A) has a major influence on accuracy, as well as a distinct influence on TCM results.
The spatial attention (S-A) improves accuracy but has almost no influence on TCM (Tab.~\ref{ablations}). 
While the temporal attention (T-A) alone can neither improve TCM and even harms accuracy, as feature aggregation may be highly noisy and imprecise without positional information~\cite{watson2021temporal}. We therefore introduce S-A with correlated 3D information serving as 3D positional encoding to ensure the temporal feature aggregation in T-A is spatially-aware, preventing accuracy degradation. 

$\mathcal{L}_{\text{m}}$ enforces consistency between the weak teacher and our prediction where they deviate significantly (i.e. $1-\mathcal{M}_{\text{m}}$), due to e.g. moving objects. When only $\mathcal{L}_{\text{m}}$ is enforced in the training (w/o geometric guidance), 
$\mathcal{M}_{\text{m}}$ can already help identify regions with large deviations due to e.g. moving objects, leading to improved accuracy~\cite{watson2021temporal}. 
From the ablations we can see $\mathcal{L}_{\text{geo}}$ is crucial for highly consistent depth predictions. The effectiveness of $\mathcal{L}_{\text{geo}}$ to respect occlusions is guaranteed with $\mathcal{M}_{\text{cycle}}$, whereas potential dynamic objects are masked by $\mathcal{M}_{\text{m}} \cdot \mathcal{M}_{\text{auto}}$. 
Hence, dynamic objects violating the static assumption of $\mathcal{L}_{\text{geo}}$ are explicitly handled with more exhaustive and accurate masking, thus the consistency performance improves.
$\mathcal{L}_{\text{geo}}$ on its own with $\mathcal{M}_{\text{min}}$ actually reduces accuracy slightly for the accuracy measure $\sigma < 1.25$ (which is in accordance with the observations from SC-SfMLearner~\cite{bian2019unsupervised}).
The additional cycle mask $\mathcal{M}_{\text{cycle}}$ can mitigate this issue by better accounting for occluded regions based on photometric cues.
$\mathcal{L}_{\text{geo}}$ together with $\mathcal{M}_{\text{cycle}}$ also significantly improves TCM. $\mathcal{L}_{\text{m}}$ further reduces the outlier rate as indicated by Sq.Rel. error, as moving objects are handled explicitly.

When spatial-temporal attention is combined with $\mathcal{L}_{\text{geo}}$ and $\mathcal{M}_{\text{cycle}}$, the additional loss function together with appropriate regularization can push the attention module to learn geometrically more consistent aggregation of temporal information by geometric guidance, thus significantly improving on TCM and depth accuracy.
The full model achieves the best results, and a larger encoder can improve results further.

Our findings are also conformed by qualitative results in Fig.~\ref{rec_ablation}. While the baseline without our contributions shows a deteriorated 3D reconstruction, the proposed geometric constraints and the novel spatial-temporal attention module both improve results individually. 

\subsection{Limitations}
Fig.~\ref{limit} (top) illustrates that the ball-query of the spatial attention can correlate spatially nearby structures. 
The temporal attention does not always provide one distinct maximum attention for the queried pixel, as multiple non-identical objects of similar appearance can correlate, yielding ambiguous attention (multiple pedestrians, multiple cars).
Only objects in a close depth layer are correlated, while other similar distant objects are ignored (e.g. cars in the background).
This behavior actually confirms our hypothesis that the spatial attention and geometry constraints guide the temporal attention towards geometry-aware aggregation of consistent features.
\setcounter{figure}{8}
\begin{figure}[!h]
      \centering
      \includegraphics[width=1.0\columnwidth]{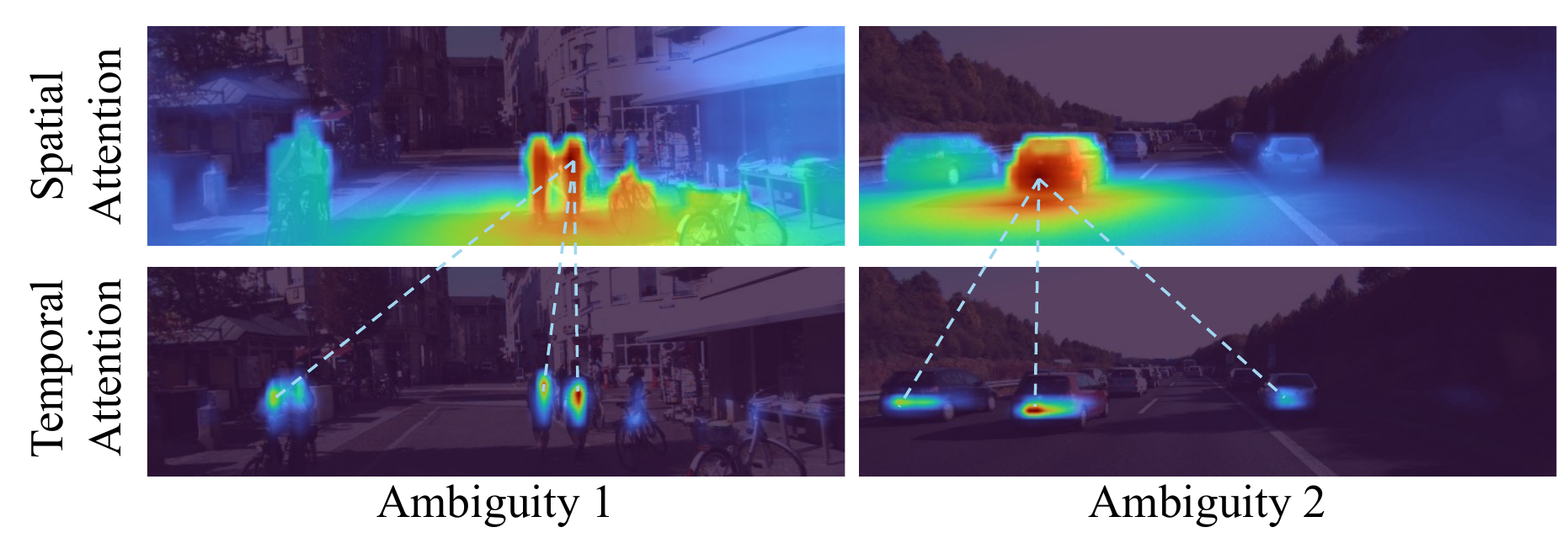}
      \caption{Illustration of spatial and temporal attention for a difficult scene with multiple similar objects.}
      \label{limit}
\end{figure}
\section{Conclusion}
\label{sec:conclusion}
To the best of our knowledge, we have for the first time presented a model that fully leverages the spatial-temporal domain to predict self-supervised consistent depth estimations by introducing a unique and novel attention model based on geometric and appearance-based information. 
Our method \textbf{TC-Depth} has proven that geometric constraints, together with cycle consistency regularization, can further improve such consistency by guiding the spatial-temporal attention aggregation. 
Future research on temporally consistent depth estimation can now be objectively compared with the new temporal consistency metric (TCM). 
\clearpage

{\small
\bibliographystyle{ieee_fullname}
\bibliography{literature}
}

\end{document}